\documentclass[a4paper, 10 pt, conference]{hsmr} 
\usepackage[utf8]{inputenc}
\IEEEoverridecommandlockouts                       
\usepackage{mathtools}

\usepackage[utf8]{inputenc}
\usepackage{mathtools}
\usepackage{geometry}
\usepackage{comment}
\usepackage[labelsep=space]{caption}
\usepackage{newtxtext,newtxmath}   
\usepackage{authblk}
\usepackage{pgfplots}
\usepackage{graphicx}
\usepackage{gensymb}
\usepackage{array}
\usepackage{stfloats}
\usepackage{soul}
\usepackage{cite}                  
\usepackage[colorlinks,urlcolor=blue]{hyperref}   
\usepackage{cleveref}              
\pgfplotsset{compat=newest} 
 
\title{\LARGE \bf
A Multi-Vine Soft Robot Enabling\\ 
Accessible Working Channel and Steering}  
\author{\\\Large Reza Kashef}
\author{\Large Cem Suulker}
\author{\Large Mohammad Sheikh Sofla}
\author{\Large Kaspar Althoefer} 
\affil{\normalsize\textit{School of Engineering and Materials Science, Queen Mary University of London, United Kingdom.}\\ \normalsize\textit{s.kasheftabrizian@qmul.ac.uk}\vspace{-0.04\linewidth}}

\begin{document}
\bstctlcite{IEEEexample:BSTcontrol}
\maketitle
\thispagestyle{empty}
\pagestyle{empty}

\section*{INTRODUCTION}

Soft eversion robots, also known as vine robots, have attracted growing interest for navigation and inspection tasks, including minimally invasive medical applications \cite{al2025tip}. A vine robot consists of a thin, flexible, inextensible tube folded inward that everts and grows forward when pressurized. This tip-growth enables navigation with minimal friction, making vine robots well suited for complex environments such as the human colon \cite{suulker2026state}.  

While their inherent softness allows passive conformation to curved pathways in confined spaces, navigation performance strongly depends on environmental interactions, including contact angle and the length of unconstrained deployed material \cite{troarxiv,haggerty2019characterizing}. Sharp directional changes, such as those in the sigmoid colon, often limit passive growth and necessitate active steering. Existing solutions include distributed artificial muscles \cite{kubler2024comparison} or dedicated tip-based steering mechanisms \cite{shi2025design}.  

In addition, many applications require payload delivery, such as sensors and tools \cite{kim2025soft,seo2024inflatable}. Within the ERC Synergy project \href{https://www.endotheranostics.com/}{EndoTheranostics}, this motivates the development of vine robots capable of delivering microsurgical tools during growth. Prior work has integrated working channels within the vine body \cite{seo2024inflatable,girerd2024material}, but these approaches constrain tool size, introduce friction, and limit access to the environment to the robot tip.  

\begin{figure}[t]
\centering
\includegraphics[width=\columnwidth]{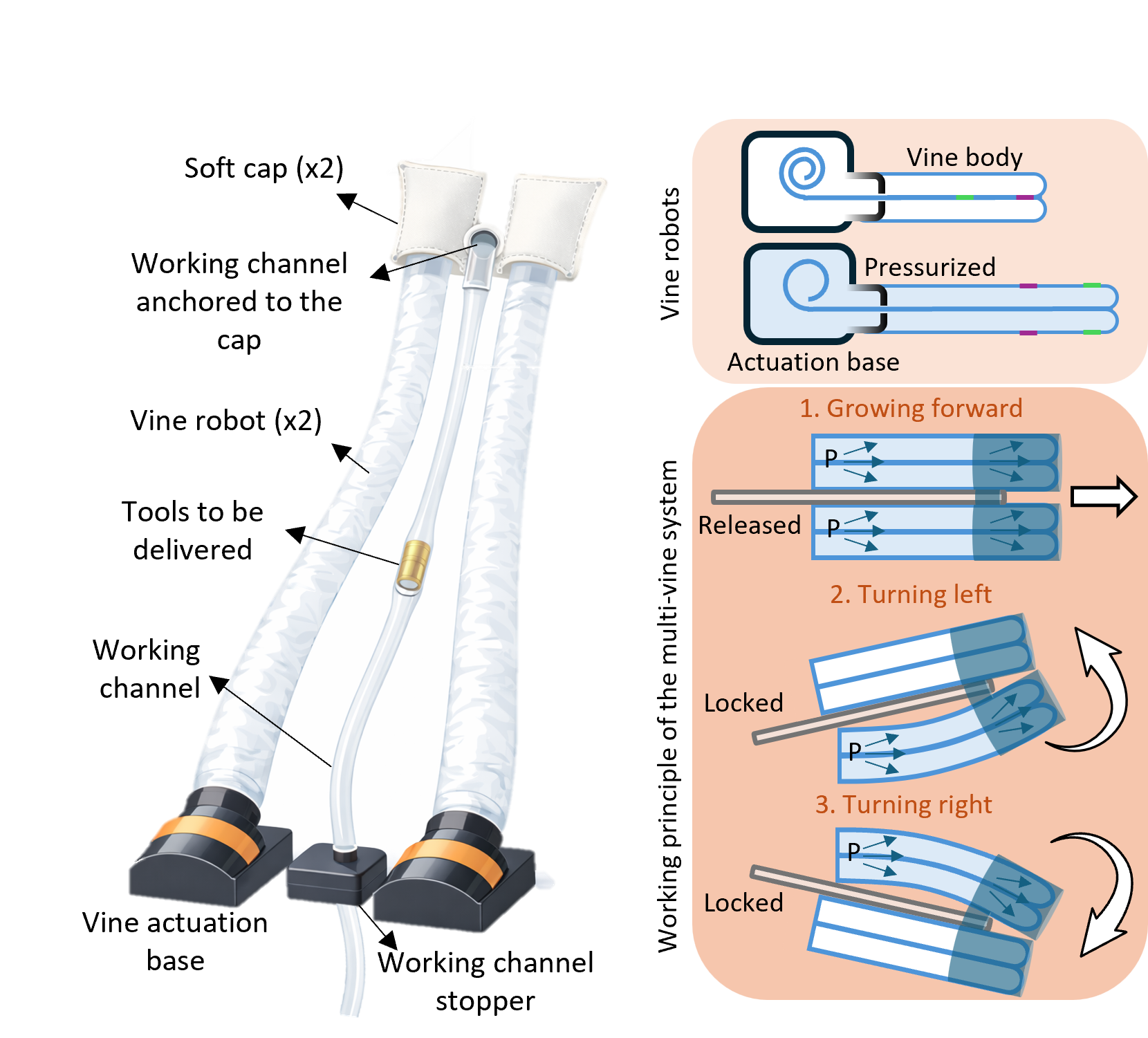}
\caption{Structure and working principle of the multi-vine system. The system consists of two vine robots equipped with soft caps that are linked together, to which an external working channel is also anchored. The left side image was assisted by Chat-GPT.}
\label{figure1}
\vspace{-0.1\linewidth}
\end{figure}

In this work, we propose a multi-vine architecture in which two vine robots are coupled to an externally integrated working channel via soft mounting tips \cite{suulker2023soft}. Independent vine actuation enables active tip steering while advancing the working channel without embedding it within the vine bodies Figure \ref{figure1}. Experiments demonstrate sharp steering of nearly 90 degrees during growth, highlighting the potential of this architecture for versatile medical and non-medical applications.

\section*{MATERIALS AND METHODS}
The robotic system grows straight by inflating both vines at the same pressure while allowing the working channel to advance freely. To achieve a left turn, the vine on the right side of the system is pressurized while the working channel is locked (Figure \ref{figure1}). This asymmetric actuation induces buckling of the structure and results in a sharp leftward turn. For turning in the opposite direction, the actuation pattern is reversed. 
The growth speed of the system can be controlled by adjusting the advancement speed of the working channel. This control strategy also ensures that the soft caps remain in their intended positions during growth. The motion settings are demonstrated in Figure \ref{figure1}.
The vine robots were fabricated from low-density polyethylene (LDPE) sheets with a thickness of 0.06 mm. The sheets were cut to the desired dimensions and ultrasonically welded using a Vetron 5064 ultrasonic welding machine to form closed-end rectangular tubes with a total length of 1.5 m and widths of 16 mm and 34 mm. Upon pressurization, the vines expanded to final diameters of approximately 10 mm and 20 mm, respectively. For colon navigation experiments, vines with a diameter of 10 mm were used. A silicone tube with an outer diameter of 6 mm was employed as the working channel. The soft caps were fabricated from a  fabric material and sewn to form closed-end cylindrical sleeves. The inner diameter of each soft cap was designed to be approximately 1 mm larger than the outer diameter of the vine, allowing the caps to slide easily over the vine tips. 
The bases of the vine robots were fabricated using filament-based 3D printing. Upon pressurization, the vine inflates, causing the inverted material, stored inside the base and wrapped around a motor-controlled spool, to evert outward, resulting in robot growth (Figure \ref{figure2}a). The pressure supply was regulated using SMC ITV2050-212BL4. The working channel advancement was controlled manually, while it is going to be motorized too.

\section*{RESULTS}
\begin{figure}[t]
\centering
\includegraphics[width=\columnwidth]{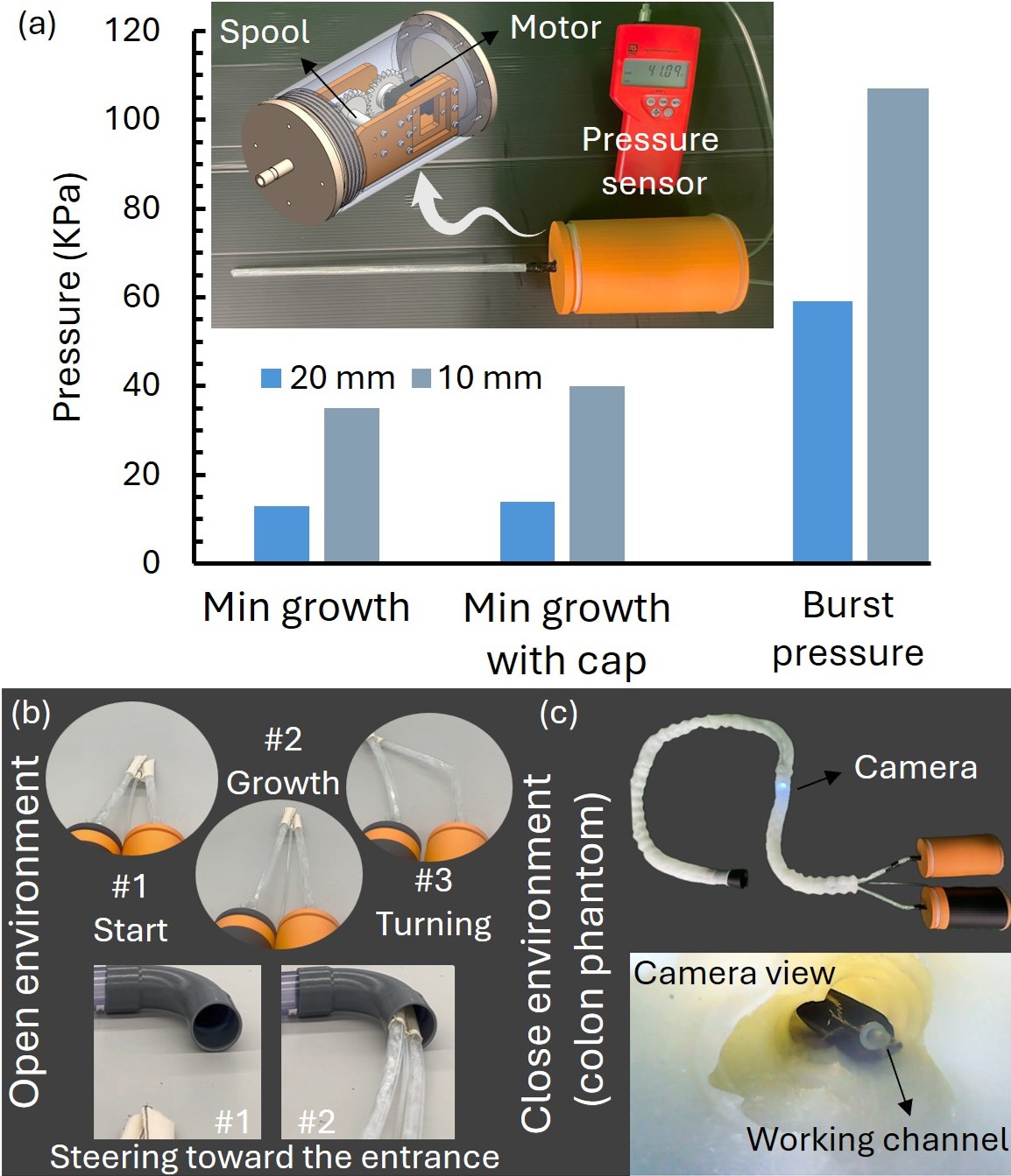}
\caption{(a) Growth pressure characterization of single vine robots with 10 mm and 20 mm diameters. (b) Navigation of the multi-vine robot in an open environment using 20 mm-diameter single vines.(c) Navigation of the multi-vine robot inside a colon phantom using 10 mm-diameter single vines.}
\label{figure2}
\vspace{-0.08\linewidth}
\end{figure}
 Figure \ref{figure2}a shows the growth pressure of single vines with diameters of 10 mm and 20 mm. As expected, due to higher friction, the smaller diameter vine requires a higher pressure to grow. However, in both cases, the growth pressure is still well below the burst pressure, indicating a sufficient safety margin during operation. Addition of the soft cap did not noticeably changed the growth pressure. The reported pressures represent the minimum required for growth; turning and steering would require higher pressures. Figure \ref{figure2}b presents the navigation of the multi-vine system in an open environment. The robot is able to grow forward and perform controlled turning. It can accurately locate and enter the opening of a pipe, demonstrating its steering capability.  Figure \ref{figure2}c shows the growth of the system inside a colon phantom with realistic colon dimensions \cite{suulker2026state}. In this test, single vines with a diameter of 10 mm were used. The phantom is made of silicone material, which is less slippery than the real colon environment. Despite this higher friction, the robot successfully passes a 90° bend while carrying the working channel along with it.


\section*{DISCUSSION}

This paper introduces a multi-vine soft robot architecture for enhanced navigation and payload delivery in vine robots. The proposed system significantly improves the navigation performance of eversion robots in both open and confined environments. By preserving an externally integrated working channel from the base to the tip, the architecture enables less restricted tool delivery, addressing a key limitation of existing vine robot designs. The multi-vine robot demonstrates sufficient maneuverability to grow through a colon phantom. While a single vine with much smaller diameter than colon, e.g. 1 cm, may grow through it, apart from lack of a working channel, its growth relies on wall contact forces that may cause discomfort.  Future work will focus on quantitatively characterizing the steering performance gains and further developing the working channel to enable tool access not only at the robot tip, but also along the entire length of the robot body.

\section*{ACKNOWLEDGMENT}
This work is supported by ERC grant EndoTheranostics, 101118626. Funded by the European Union. Views and opinions expressed are however those of the author(s) only and do not necessarily reflect those of the European Union or the European Research Council Executive Agency. Neither the European Union nor the granting authority can be held responsible for them.



\nocite{*}
\bibliographystyle{IEEEtran}
\bibliography{HSMR}

\end{document}